# Random Algorithms for the Loop Cutset Problem

Ann Becker, Reuven Bar-Yehuda, Dan Geiger*
Computer Science Department
Technion, ISRAEL
{anyuta,reuven,dang}@cs.technion.ac.il

## Abstract

We show how to find a minimum loop cutset in a Bayesian network with high probability. Finding such a loop cutset is the first step in Pearl's method of conditioning for inference. Our random algorithm for finding a loop cutset, called REPEATEDWGUESSI, outputs a minimum loop cutset, after $O(c \cdot 6^k kn)$ steps, with probability at least $1-(1-\frac{1}{6^k})^{c6^k}$, where $c > 1$ is a constant specified by the user, $k$ is the size of a minimum weight loop cutset, and $n$ is the number of vertices. We also show empirically that a variant of this algorithm, called WRA, often finds a loop cutset that is closer to the minimum loop cutset than the ones found by the best deterministic algorithms known.

## 1 Introduction

All exact inference algorithms for the computation of a posterior probability in general Bayesian networks have two conceptual phases. One phase handles operations on the graphical structure itself and the other performs probabilistic computations. For example, the clique tree algorithm requires us to first find a "good" clique tree and then perform probabilistic computations on the clique tree [LS88]. Pearl's method of conditioning requires us first to find a "good" loop cutset and then perform a calculation for each loop cutset [Pe86, Pe88]. Finally, Shachter's algorithm requires us to find a "good" sequence of transformations and then, for each transformation, to compute some conditional probability tables [Sh86].

In the three algorithms just mentioned the first phase is to find a good discrete structure, namely, a clique tree, a cutset, or a sequence of transformations. The goodness of the structure depends on a chosen parameter that, if selected appropriately, reduces the probabilistic computations done in the second phase. Finding a structure that optimizes the selected parameter is usually NP-hard and thus heuristic methods are applied to find a reasonable structure. Most methods in the past had no guarantee of performance and performed very badly when presented with an appropriate example. Becker and Geiger offered an algorithm that finds a loop cutset for which the logarithm of the state space is guaranteed to be a constant factor off the optimal value [BG94, BG96]. Bafna et al. and Fujito developed similar algorithms [BBF95, Fu96].

While the approximation algorithms for the loop cutset problem are quite useful, it is worthwhile to invest in finding a minimum loop cutset, rather than an approximation, because the cost of finding such a loop cutset is amortized over the many iterations of the conditioning method. In fact, one may invest an effort of complexity exponential in the size of the loop cutset or even larger in finding a minimum loop cutset because the second phase of the conditioning algorithm, which is repeated for many iterations, uses a procedure of such complexity. The same considerations apply also to constraint satisfaction problems [De90].

In this paper we describe several random algorithms that compute a loop cutset. As in [BGNR94], our solution is based on a reduction to the *Weighted Feedback Vertex Set (WFVS) Problem*, defined below. A feedback vertex set $F$ is a set of vertices of an undirected graph $G = (V, E)$ such that by removing $F$ from $G$, along with all the edges incident with $F$, a set of trees is obtained. The weighted feedback vertex set (WFVS) problem is to find a feedback vertex set $F$ of a vertex-weighted graph $(G, w)$, $w : V \to \mathbb{R}^+$, such that $\sum_{v \in F} w(v)$ is minimized. When $w(v) \equiv 1$, this problem is called the FVS problem. The decision version associated with the FVS problem is known to be NP-Complete [GJ79, pp. 191-192]. Note that also the problem of finding a minimum loop cutset is NP-complete [SC90].

Our random algorithm for finding a FVS, called REPEATEDWGUESSI, outputs a minimum weight FVS, after $O(c \cdot 6^k kn)$ steps, with probability at least $1 - $

---

$(1 - \frac{1}{6^k})^{c6^k}$, where $c > 1$ is a constant specified by the user, $k$ is the size of a minimum weight FVS, and $n$ is the number of vertices. For unweighted graphs we present an algorithm that finds a minimum FVS of a graph $G$, after $O(4^k kn)$ steps, with probability at least $1 - (1 - \frac{1}{4^k})^{c4^k}$. In comparison, several deterministic algorithms for finding a minimum FVS are described in the literature. One has a complexity $O((2k+1)^k n^2)$ [DF95] and others have a complexity $O((17k^4)!n)$ [Bo90, DF92].

A final variant of our random algorithms, called WRA, has the best performance because it utilizes information from previous runs. This algorithm is harder to analyze and its investigation is mostly experimental. We show empirically that the actual run time of WRA is comparable to a Modified Greedy Algorithm (MGA), devised by Becker and Geiger [BG96], which is the best available deterministic algorithm for finding close to optimal loop cutsets, and yet, the output of WRA is often closer to the minimum loop cutest than the output of MGA.

The rest of the paper is organized as follows. In Section 2 we outline the method of conditioning, explain the related loop cutset problem and describe the reduction from the loop cutset problem to the Weighted Feedback Vertex Set (WFVS) Problem. In Section 3 we present three random algorithms for the WFVS problem and their analysis. In Section 4 we compare experimentally WRA and MGA wrt performance and run time.

## 2  Background: The loop cutset problem

Pearl's method of conditioning is one of the best known inference methods for Bayesian networks. It is a method of choice in some genetic linkage programs [Ot91, BGS98]. A short overview of the method of conditioning, and definitions related to Bayesian networks, are given below. See [Pe88] for more details. We then define the loop cutset problem.

Let $P(u_1, \ldots, u_n)$ be a probability distribution where each $u_i$ draws values from a finite set called the *domain* of $u_i$. A directed graph $D$ with no directed cycles is called a *Bayesian network of* $P$ if there is a 1-1 mapping between $\{u_1, \ldots, u_n\}$ and vertices in $D$, such that $u_i$ is associated with vertex $i$ and $P$ can be written as follows:

$$P(u_1, \ldots, u_n) = \prod_{i=1}^{n} P(u_i \mid u_{i_1}, \ldots, u_{i_{j(i)}}) \quad (1)$$

where $i_1, \ldots, i_{j(i)}$ are the source vertices of the incoming edges to vertex $i$ in $D$.

Suppose now that some variables $\{v_1, \ldots, v_l\}$ among $\{u_1, \ldots, u_n\}$ are assigned specific values $\{\mathbf{v}_1, \ldots, \mathbf{v}_l\}$ respectively. The *updating problem* is to compute the probability $P(u_i \mid v_1 = \mathbf{v}_1, \ldots, v_l = \mathbf{v}_l)$ for $i = 1, \ldots, n$.

A *trail* in a Bayesian network is a subgraph whose underlying graph is a simple path. A vertex $b$ is called a *sink* with respect to a trail $t$ if there exist two consecutive edges $a \to b$ and $b \leftarrow c$ on $t$. A trail $t$ is *active by a set of vertices* $Z$ if (1) every sink with respect to $t$ either is in $Z$ or has a descendant in $Z$ and (2) every other vertex along $t$ is outside $Z$. Otherwise, the trail is said to be *blocked (d-separated)* by $Z$.

Verma and Pearl [VP88] have proved that if $D$ is a Bayesian network of $P(u_1, \ldots, u_n)$ and all trails between a vertex in $\{r_1, \ldots, r_l\}$ and a vertex in $\{s_1, \ldots, s_k\}$ are blocked by $\{t_1, \ldots, t_m\}$, then the corresponding sets of variables $\{u_{r_1}, \ldots, u_{r_l}\}$ and $\{u_{s_1}, \ldots, u_{s_k}\}$ are independent conditioned on $\{u_{t_1}, \ldots, u_{t_m}\}$. Furthermore, Geiger and Pearl proved a converse to this theorem [GP90]. Both results are presented and extended in [GVP90].

Using the close relationship between blocked trails and conditional independence, Kim and Pearl [KP83] developed an algorithm UPDATE-TREE that solves the updating problem on Bayesian networks in which every two vertices are connected with at most one trail (singly-connected). Pearl then solved the updating problem on any Bayesian network as follows [Pe86]. First, a set of vertices $S$ is selected such that any two vertices in the network are connected by at most one *active* trail in $S \cup Z$, where $Z$ is any subset of vertices. Then, UPDATE-TREE is applied once for each combination of value assignments to the variables corresponding to $S$, and, finally, the results are combined. This algorithm is called the method of *conditioning* and its complexity grows exponentially with the size of $S$. The set $S$ is called a *loop cutset*. Note that when the domain size of the variables varies, then UPDATE-TREE is called a number of times equal to the product of the domain sizes of the variables whose corresponding vertices participate in the loop cutset. If we take the logarithm of the domain size (number of values) as the weight of a vertex, then finding a loop cutset such that the sum of its vertices weights is minimum optimizes Pearl's updating algorithm in the case where the domain sizes may vary.

We now give an alternative definition for a loop cutset $S$ and then provide a probabilistic algorithm for finding it. This definition is borrowed from [BGNR94]. The underlying graph $G$ of a directed graph $D$ is the undirected graph formed by ignoring the directions of the edges in $D$. A *cycle* in $G$ is a path whose two terminal vertices coincide. A *loop* in $D$ is a subgraph of $D$ whose underlying graph is a cycle. A vertex $v$ is a *sink* with respect to a loop $\Gamma$ if the two edges adjacent to $v$ in $\Gamma$ are directed into $v$. Every loop must contain at least one vertex that is not a sink with respect to that loop. Each vertex that is not a sink with respect to a loop $\Gamma$ is called an *allowed vertex with re-* spect to $\Gamma$. A *loop cutset* of a directed graph $D$ is a set of vertices that contains at least one allowed vertex with respect to each loop in $D$. The weight of a set of vertices $X$ is denoted by $w(X)$ and is equal to $\sum_{v \in X} w(v)$ where $w(x) = \log(|x|)$ and $|x|$ is the size of the domain associated with vertex $x$. A *minimum loop cutset* of a weighted directed graph $D$ is a loop cutset $F^*$ of $D$ for which $w(F^*)$ is minimum over all loop cutsets of $G$. The *Loop Cutset Problem* is defined as finding a minimum loop cutset of a given weighted directed graph $D$.

The approach we take is to reduce the weighted loop cutset problem to the weighted feedback vertex set problem, as done by [BGNR94]. We now define the weighted feedback vertex set problem and then the reduction.

Let $G = (V, E)$ be an undirected graph, and let $w : V \to \mathbb{R}^+$ be a weight function on the vertices of $G$. A *feedback vertex set* of $G$ is a subset of vertices $F \subseteq V$ such that each cycle in $G$ passes through at least one vertex in $F$. In other words, a feedback vertex set $F$ is a set of vertices of $G$ such that by removing $F$ from $G$, along with all the edges incident with $F$, we obtain a set of trees (i.e., a forest). The weight of a set of vertices $X$ is denoted (as before) by $w(X)$ and is equal to $\sum_{v \in X} w(v)$. A *minimum feedback vertex set* of a weighted graph $G$ with a weight function $w$ is a feedback vertex set $F^*$ of $G$ for which $w(F^*)$ is minimum over all feedback vertex sets of $G$. The *Weighted Feedback Vertex Set (WFVS) Problem* is defined as finding a minimum feedback vertex set of a given weighted graph $G$ having a weight function $w$.

The reduction is as follows. Given a weighted directed graph $(D, w)$ (e.g., a Bayesian network), we define the *splitting* weighted undirected graph $D_s$ with a weight function $w_s$ as follows. Split each vertex $v$ in $D$ into two vertices $v_{\text{in}}$ and $v_{\text{out}}$ in $D_s$ such that all incoming edges to $v$ in $D$ become undirected incident edges with $v_{\text{in}}$ in $D_s$, and all outgoing edges from $v$ in $D$ become undirected incident edges with $v_{\text{out}}$ in $D_s$. In addition, connect $v_{\text{in}}$ and $v_{\text{out}}$ in $D_s$ by an undirected edge. Now set $w_s(v_{\text{in}}) = \infty$ and $w_s(v_{\text{out}}) = w(v)$. For a set of vertices $X$ in $D_s$, we define $\psi(X)$ as the set obtained by replacing each vertex $v_{\text{in}}$ or $v_{\text{out}}$ in $X$ by the respective vertex $v$ in $D$ from which these vertices originated.

Our algorithm can now be easily stated.

**Algorithm RLC**

**Input:** *A Bayesian network $D$*
**Output:** *A loop cutset of $D$*
  1. Construct the splitting graph $D_s$
     with weight function $w_s$;
  2. Apply WRA on $(D_s, w_s)$ to obtain
     a feedback vertex set $F$;
  3. Output $\psi(F)$.

It is immediately seen that if WRA outputs a feedback vertex set $F$ of $D_s$ whose weight is minimum with high probability, then $\psi(F)$ is a loop cutset of $D$ with minimum weight with the same probability. This observation holds because there is an obvious one-to-one and onto correspondence between loops in $D$ and cycles in $D_s$ and because WRA never chooses a vertex that has an infinite weight.

## 3 Algorithms for the WFVS problem

Recall that a feedback vertex set of $G$ is a subset of vertices $F \subseteq V$ such that each cycle in $G$ passes through at least one vertex in $F$. In Section 3.1 we address the problem of finding a FVS with a minimum number of vertices and in Sections 3.2 and 3.3 we address the problem of finding a FVS with a minimum weight. Throughout, we allow $G$ to have parallel edges. If two vertices $u$ and $v$ have parallel edges between them, then every FVS of $G$ includes either $u$, $v$, or both.

### 3.1 The basic algorithms

In this section we present a random algorithm for the FVS problem. First we introduce some additional terminology and notation. Let $G = (V, E)$ be an undirected graph. The degree of a vertex $v$ in $G$, denoted by $d(v)$, is the number of vertices adjacent to $v$. A *self-loop* is an edge with two endpoints at the same vertex. A *leaf* is a vertex with degree less or equal 1, a *linkpoint* is a vertex with degree 2 and a *branchpoint* is a vertex with degree strictly higher than 2. The cardinality of a set $X$ is denoted by $|X|$.

A graph is called *rich* if every vertex is a branchpoint and it has no self-loops. Given a graph $G$, by repeatedly removing all leaves, and bypassing every linkpoint with an edge, a graph $G'$ is obtained such that the size of a minimum FVS in $G'$ and in $G$ are equal and every minimum FVS of $G'$ is a minimum WFVS of $G$. Since every vertex involved in a self-loop belongs to every FVS, we can transform $G'$ to a rich graph $G^r$ by adding the vertices involved in self loops to the output of the algorithm.

Our algorithm is based on the observation that if we pick an edge at random from a rich graph there is a probability of at least 1/2 that at least one endpoint of the edge belongs to any given FVS $F$. A precise formulation of this claim is given by Lemma 1 whose proof appears implicitly in [Vo68, Lemma 4].

**Lemma 1** *Let $G = (V, E)$ be a rich graph, $F$ be a feedback vertex set of $G$ and $X = V \setminus F$. Let $E_X$ denote the set of edges in $E$ whose endpoints are all vertices in $X$ and $E_{F,X}$ denote the set of edges in $G$ that connect vertices in $F$ with vertices in $X$. Then, $|E_X| \leq |E_{F,X}|$.*

**Proof.**    The graph obtained by deleting a feedback vertex set $F$ of a graph $G(V, E)$ is a forest with vertices $X = V \setminus F$. Hence, $|E_X| < |X|$. However, each vertex in $X$ is a branchpoint in $G$, and so,

$$3|X| \leq \sum_{v \in X} d(v) = |E_{F,X}| + 2|E_X|.$$

Thus, $|E_X| \leq |E_{F,X}|$. □

Lemma 1 implies that when picking an edge at random from a rich graph, it is at least as likely to pick an edge in $E_{F,X}$ than an edge in $E_X$. Consequently, selecting a vertex at random from a randomly selected edge has a probability of at least 1/4 to belong to a minimum FVS. This idea yields a simple algorithm to find a FVS.

## ALGORITHM SingleGuess(G,k)

**Input:** *An undirected graph $G_0$ and an integer $k > 0$.*
**Output:** *A feedback vertex set $F$ of size $\leq k$, or "Fail" otherwise.*
    For $i = 1, \ldots, k$
       1. Reduce $G_{i-1}$ to a rich graph $G_i$ while placing self loop vertices in $F$.
       2. If $G_i$ is the empty graph **Return** $F$
       3. Pick an edge $e = (u, v)$ at random from $E_i$
       4. Pick a vertex $v_i$ at random from $(u, v)$
       5. $F \leftarrow F \cup \{v_i\}$
       6. $V \leftarrow V \setminus \{v_i\}$
    **Return** "Fail"

Due to Lemma 1, when SINGLEGUESS terminates with a FVS of size $k$, there is a probability of at least $1/4^k$ that the output is a minimum FVS.

Note that steps 1 and 2 in SINGLEGUESS determine a vertex $v$ by first selecting an arbitrary edge and then selecting an arbitrary endpoint of this edge. An equivalent way of achieving the same selection rule is to choose a vertex with probability proportional to its degree:

$$p(v) = \frac{d(v)}{\sum_{u \in V} d(u)} = \frac{d(v)}{2 \cdot |E|}$$

To see the equivalence of these two selection methods, define $\Gamma(v)$ to be a set of edges whose one endpoint is $v$, and note that for graphs without self-loops,

$$p(v) = \sum_{e \in \Gamma(v)} p(v|e) \cdot p(e) = \frac{1}{2} \sum_{e \in \Gamma(v)} p(e) = \frac{d(v)}{2 \cdot |E|}$$

This equivalent phrasing of the selection criterion is easier to extend to the weighted case and will be used in the following sections.

An algorithm for finding a minimum FVS with high probability, which we call REPEATEDGUESS, can now be described as follows: Start with $k = 1$. Repeat SINGLEGUESS $c \cdot 4^k$ times where $c > 1$ is a parameter defined by the user. If in one of the iterations a FVS of size $\leq k$ is found, then output this FVS, otherwise, increase $k$ by one and continue.

## ALGORITHM RepeatedGuess(G,c)

**Input:** *An undirected graph $G$ and a constant $c > 1$.*
**Output:** *A feedback vertex set $F$.*
    For $k = 1, \ldots, |V|$
       **Repeat** $c \cdot 4^k$ **times**
           1. $F \leftarrow$ SINGLEGUESS$(G, k)$
           2. If $F$ is not "Fail" **then Return** $F$
       **End** {Repeat}
    **End** {For}

The main claims about these algorithms are given by the following theorem.

**Theorem 2** *Let $G$ be an undirected graph and $c \geq 1$ be a constant. Then,* SINGLEGUESS$(G, k)$ *outputs a FVS whose expected size is no more than $4k$, and* REPEATEDGUESS$(G, c)$ *outputs, after $O(4^k kn)$ steps, a minimum FVS with probability at least $1 - (1 - \frac{1}{4^k})^{c4^k}$, where $k$ is the size of a minimum FVS and $n$ is the number of vertices.*

The claims about the probability of success and number of steps follow immediately from the fact that the probability of success of SingleGuess is at least $(1/4)^k$ and that, in case of success, $O(4^k)$ iterations are performed each taking $O(kn)$ steps. The proof about the expected size of a single guess is presented in the next section.

Theorem 2 shows that each guess produces a FVS which, on the average, is not too far from the minimum, and that after enough iterations, the algorithm converges to the minimum with high probability. In the weighted case, discussed next, we managed to achieve each of these two guarantees in a separate algorithm, but we were unable to achieve both guarantees in a single algorithm.

### 3.2 The weighted algorithms

We now turn to the weighted FVS problem (WFVS) of size $k$ which is to find a feedback vertex set $F$ of a vertex-weighted graph $(G, w)$, $w : V \to \mathbb{R}^+$, of size less or equal $k$ such that $w(F)$ is minimized.

Note that for the weighted FVS problem we cannot replace each linkpoint $v$ with an edge, since if $v$ has weight lighter than its branchpoint neighbors then $v$ can participate in a minimum weight FVS of size $k$.

A graph is called *branchy* if it has no endpoints, no self loops, and, in addition, each linkpoint is connected only to branchpoints [BGNR94]. Given a graph $G$, by repeatedly removing all leaves, and bypassing with an edge every linkpoint that has a neighbor with equal or lighter weight, a graph $G'$ is obtained such that the weight of a minimum weight FVS (of size $k$) in $G'$ and in $G$ are equal and every minimum WFVS of $G'$ is a minimum WFVS of $G$. Since every vertex with a self-loop belongs to every FVS, we can transform $G'$ to a branchy graph without self-loops by adding the vertices involved in self loops to the output of the algorithm.

To address the WFVS problem we offer two slight modifications to the algorithm SINGLEGUESS presented in the previous section. The first algorithm, which we call SINGLEWGUESSI, is identical to SINGLEGUESS except that in each iteration we make a reduction to a branchy graph instead of a reduction to a rich graph. It chooses a vertex with probability proportional to the degree using $p(v) = d(v)/\sum_{u \in V} d(u)$. Note that this probability does not take the weight of a vertex into account. A second algorithm, which we call SINGLEWGUESSII, chooses a vertex with probability proportional to the ratio of its degree over its weight,

$$p(v) = \frac{d(v)}{w(v)} / \sum_{u \in V} \frac{d(u)}{w(u)}. \quad (2)$$

**ALGORITHM SingleWGuessI(G,k)**

Input: An undirected weighted graph $G_0$ and an integer $k > 0$.
Output: A feedback vertex set $F$ of size $\leq k$, or "Fail" otherwise.
  For $i = 1, \ldots, k$
    1. Reduce $G_{i-1}$ to a branchy graph $G_i(V_i, E_i)$ while placing self loop vertices in $F$.
    2. If $G_i$ is the empty graph **Return** $F$
    3. Pick a vertex $v_i \in V_i$ at random with probability $p_i(v) = d_i(v)/\sum_{u \in V_i} d_i(u)$
    4. $F \leftarrow F \cup \{v_i\}$
    5. $V \leftarrow V \setminus \{v_i\}$
  **Return** "Fail"

The second algorithm uses Eq 2 for computing $p(v)$ in Line 1. These two algorithms have remarkably different guarantees of performance. Version I guarantees that choosing a vertex that belongs to any given FVS is larger than 1/6, however, the expected weight of a FVS produced by version I cannot be bounded by a constant times the weight of a minimum WFVS. Version II guarantees that the expected weight of its output is bounded by 6 times the weight of a minimum WFVS, however, the probability of converging to a minimum after any fixed number of iterations can be arbitrarily small. We first demonstrate via an example the negative claims. The positive claims are phrased more precisely in Theorem 3 and proven thereafter.

Consider the graph shown in Figure 1 with three vertices $a,b$ and $c$, and corresponding weights $w(a) = 6$, $w(b) = 3\epsilon$ and $w(c) = 3m$, with three parallel edges between $a$ and $b$, and three parallel edges between $b$ and $c$. The minimum WFVS $F^*$ with size 1 consists of vertex $a$. According to Version II, the probability of choosing vertex $a$ is (Eq. 2):

$$p(a) = \frac{\epsilon}{(1 + 1/m) \cdot \epsilon + 1}$$

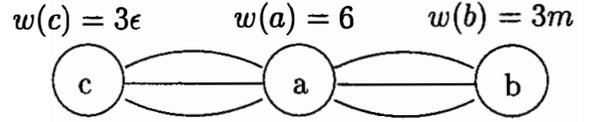

Figure 1: The minimum WFVS $F^* = \{a\}$.

So if $\epsilon$ is arbitrarily small and $m$ is sufficiently large, then the probability of choosing vertex $a$ is arbitrarily small. Thus, the probability of choosing a vertex from some $F^*$ by the criterion $d(v)/w(v)$, as done by Version II, can be arbitrarily small. If, on the other hand, Version I is used, then the probability of choosing $a, b$, or $c$ is $1/2, 1/4, 1/4$, respectively. Thus, the expected weight of the first vertex to be chosen is $3/4 \cdot (\epsilon + m + 4)$, while the weight of a minimum WFVS is 6. Consequently, if $m$ is sufficiently large, the expected weight of a WFVS found by Version I can be arbitrarily larger than a minimum WFVS.

The algorithm for repeated guesses, which we call REPEATEDWGUESSI is as follows: repeat SINGLEWGUESSI $c \cdot 6^k$ times, where $k$ is the number of vertices (size) of a minimum WFVS we seek. If no FVS is found of size $\leq k$, the algorithm outputs that the size of a minimum WFVS is larger than $k$ with high probability, otherwise, it outputs the lightest FVS of size less or equal $k$ among those explored. The following theorem summarizes the main claims.

**Theorem 3** Let $G$ be a weighted undirected graph and $c \geq 1$ be a constant.
a) The algorithm REPEATEDWGUESSI$(G, c)$ outputs, after $O(6^k kn)$ steps, a minimum FVS with probability at least $1 - (1 - \frac{1}{6^k})^{c6^k}$, where $k$ is the size of a minimum weight FVS of $G$ and $n$ is the number of vertices.
b) The algorithm SINGLEWGUESSII$(G)$ outputs a feedback vertex set whose expected weight is no more than six times the weight of the minimum WFVS.

The proof of each part requires a preliminary lemma.

**Lemma 4** Let $G = (V, E)$ be a branchy graph, $F$ be a feedback vertex set of $G$ and $X = V \setminus F$. Let $E_X$ denote the set of edges in $E$ whose endpoints are all vertices in $X$ and $E_{F,X}$ denote the set of edges in $G$ that connect vertices in $F$ with vertices in $X$. Then, $|E_X| \leq 2 \cdot |E_{F,X}|$.

**Proof.**  Let $X^b$ be the set of branchpoints in $X$. We replace every linkpoint in $X$ by an edge between its neighbors, and denote the resulting set of edges between vertices in $X^b$ by $E^b_{X^b}$ and between vertices in $X^b$ and $F$ by $E^b_{F,X^b}$. The proof of Lemma 1 shows that

$$|E^b_{X^b}| \leq |E^b_{F,X^b}|.$$

Since every linkpoint in $X$ has both neighbors in the set $X^b \cup F$, the following holds:

$$|E_X| \leq 2 \cdot |E^b_{X^b}| \text{ and } |E_{F,X}| = |E^b_{F,X^b}|.$$

Hence, $|E_X| \leq 2 \cdot |E_{F,X}|$. □

An immediate consequence of Lemma 4 is that the probability of randomly choosing an edge that has at least one endpoint that belongs to a FVS is greater or equal 1/3. Thus, selecting a vertex at random from a randomly selected edge has a probability of at least 1/6 to belong to a FVS. Consequently, if the algorithm terminates after $c \cdot 6^k$ iterations, with a WFVS of size $k$, there is a probability of at least $1 - (1 - \frac{1}{6^k})^{c6^k}$ that the output is a minimum WFVS of size at most $k$. This proves part (a) of Theorem 3.

The second part requires the following lemma.

**Lemma 5** *Let $G$ be a branchy graph and $F$ be a FVS of $G$. Then,*

$$\sum_{v \in V} d(v) \leq 6 \sum_{v \in F} d(v).$$

**Proof.** Denote by $d_Y(v)$ the number of edges between a vertex $v$ and a set of vertices $Y$. Then,

$$\sum_{v \in V} d(v) = \sum_{v \in X} d(v) + \sum_{v \in F} d(v) =$$
$$\sum_{v \in X} d_X(v) + \sum_{v \in X} d_F(v) + \sum_{v \in F} d(v).$$

Due to Lemma 4,

$$\sum_{v \in X} d_X(v) = 2|E_X| \leq 4|E_{F,X}| = 4 \sum_{v \in X} d_F(v). \quad (3)$$

Consequently,

$$\sum_{v \in V} d(v) \leq 4 \sum_{v \in X} d_F(v) +$$
$$\sum_{v \in X} d_F(v) + \sum_{v \in F} d(v) \leq 6 \sum_{v \in F} d(v)$$

as claimed. □

We can now prove part (b) of Theorem 3 analyzing the performance of SINGLEWGUESSII(G). Recall that $V_i$ is the set of vertices in graph $G_i$ in iteration $i$, $d_i(v)$ is the degree of vertex $v$ in $G_i$, and $v_i$ is the vertex chosen in iteration $i$. Furthermore, recall that $p_i(v)$ is the probability to choose vertex $v$ in iteration $i$.

The expected weight $E_i(w(v)) = \sum_{v \in V_i} w(v) \cdot p_i(v)$ of a chosen vertex in iteration $i$ is denoted with $a_i$. Thus, due to the linearity of the expectation operator, $E(w(F)) = \sum_{i=1}^{k} a_i$, assuming $|F| = k$. We define a normalization constant for iteration $i$ as follows:

$$\gamma_i = \left[ \sum_{u \in V_i} \frac{d_i(u)}{w(u)} \right]^{-1}$$

Then, $p_i(v) = \gamma_i \cdot \frac{d_i(v)}{w(v)}$ and

$$a_i = \sum_{v \in V_i} w(v) \cdot \frac{d_i(v)}{w(v)} \cdot \gamma_i = \gamma_i \cdot \sum_{v \in V_i} d_i(v)$$

Let $F^*$ be a minimum FVS of $G$ and $F^*_i$ be minimum weight FVS of the graph $G_i$. The expected weight $E_i(w(v)|v \in F^*_i))$ of a vertex chosen from $F^*_i$ in iteration $i$ is denoted with $b_i$. We have,

$$b_i = \sum_{v \in F^*_i} w(v) \cdot p_i(v) = \gamma_i \cdot \sum_{v \in F^*_i} d_i(v)$$

By Lemma 5, $a_i/b_i \leq 6$ for every $i$.

Recall that by definition $F^*_2$ is the minimum FVS in the branchy graph $G_2$ obtained from $G_1 \setminus \{v_1\}$. We get,

$$E(w(F^*)) \geq E_1(w(v)|v \in F^*_1)) + E(w(F^*_2))$$

because the right hand side is the expected weight of the output $F$ assuming the algorithm finds a minimum FVS on $G_2$ and just needs to select one additional vertex, while the left hand side is the unrestricted expectation. By repeating this argument we get,

$$E(w(F^*)) \geq b_1 + E(w(F^*_2)) \geq \cdots \geq \sum_{i=1}^{k} b_i$$

Using $\sum_i a_i / \sum_i b_i \leq \max_i a_i/b_i \leq 6$, we obtain

$$E(w(F)) \leq 6 \cdot E(w(F^*)).$$

Hence, $E(w(F)) \leq 6 \cdot w(F^*)$ as claimed. □

The proof that SINGLEGUESS$(G, k)$ outputs a FVS whose expected size is no more than $4k$ (Theorem 2) where $k$ is the size of a minimum FVS is analogous to the proof of Theorem 3 in the following sense. We assign a weight 1 to all vertices and replace the reference to Lemma 5 by a reference to the following claim: If $F$ is a FVS of a rich graph $G$, then $\sum_{v \in V} d(v) \leq 4 \sum_{v \in F} d(v)$. The proof of this claim is identical to the proof of Lemma 5 except that instead of using Lemma 4 we use Lemma 1.

### 3.3 The practical algorithm

In previous sections we presented several algorithms for finding minimum FVS with high probability. The description of these algorithms was geared towards analysis, rather than as a prescription to a programmer. In particular, SINGLEWGUESSI(G,K) discards all the work done for finding a FVS whenever more than $k$ vertices are chosen. This feature allowed us to regard each call to SINGLEWGUESSI(G,K) made by REPEATEDWGUESSI as an independent process. Furthermore, there is a small probability for a very long run even when the size of the minimum FVS is small.

We now slightly modify REPEATEDWGUESSI to obtain an algorithm, termed WRA, which does not suffer from these deficiencies. The new algorithm works as follows. Repeat SINGLEWGUESSI$(G, |V|)$ for $min(Max, c \cdot 6^{w(F)})$ iterations, where $w(F)$ is the weight of the lightest WFVS found so far and $Max$ is some specified constant determining the maximum number of iterations of SINGLEWGUESSI.

**ALGORITHM WRA$(G, c, Max)$**

**Input:** An undirected weighted graph $G(V, E)$ and constants $Max$ and $c > 1$
**Output:** A feedback vertex set $F$
  $F \leftarrow$ SINGLEWGUESSI $(G, |V|)$
  $M \leftarrow min(Max, c \cdot 6^{w(F)})$
  $i \leftarrow 1$;
  **While** $i \leq M$ **do**
    1. $F' \leftarrow$ SINGLEWGUESSI$(G, |V|)$
    2. **If** $w(F') \leq w(F)$ **then**
        $F \leftarrow F'$;
        $M \leftarrow min(Max, c \cdot 6^{w(F)})$
    3. $i \leftarrow i + 1$;
  **End** {While}
  Return $F$

**Theorem 6** *If $Max \geq c6^k$, where $k$ is the size of a minimum WFVS of an undirected weighted graph $G$, then WRA$(G, c, Max)$ outputs a minimum WFVS of $G$ with probability at least $1 - (1 - \frac{1}{6^k})^{c6^k}$.*

The proof is an immediate corollary of Theorem 3.

The choise of $Max$ and $c$ depend on the application. A decision-theoretic approach for selecting such values for any-time algorithms is discussed in [BH90].

## 4  Experimental results

The experiments compared the outputs of WRA vis-à-vis a greedy algorithm GA and a modified greedy algorithm MGA [BG96] based on randomly generated graphs and on some real graphs contributed by the Hugin group (www.hugin.com).

The random graphs are divided into three sets. Graphs with 15 vertices and 25 edges where the number of values associated with each vertex is randomly chosen between 2 and 6, 2 and 8, and between 2 and 10.

| $|V|$ | $|E|$ | values | size | MGA | WRA | Eq. |
|---|---|---|---|---|---|---|
| 15 | 25 | 2–6 | 3–6 | 12 | 81 | 7 |
| 15 | 25 | 2–8 | 3–6 | 7 | 89 | 4 |
| 15 | 25 | 2–10 | 3–6 | 6 | 90 | 4 |
| 25 | 55 | 2–6 | 7–12 | 3 | 95 | 2 |
| 25 | 55 | 2–8 | 7–12 | 3 | 97 | 0 |
| 25 | 55 | 2–10 | 7–12 | 0 | 100 | 0 |
| 55 | 125 | 2–10 | 17–22 | 0 | 100 | 0 |
|  |  |  |  | 31 | 652 | 17 |

Figure 2: Number of graphs in which MGA or WRA yield a smaller loop cutset. Each line is based on 100 graphs.

Graphs with 25 vertices and 55 edges where the number of values associated with each vertex is randomly chosen between 2 and 6, 2 and 8, and between 2 and 10. Graphs with 55 vertices and 125 edges where the number of values associated with each vertex is randomly chosen between 2 and 10. Each instance of the three classes is based on 100 random graphs generated as described by [SC90]. The total number of random graphs we used is 700.

The results are summarized in the table below. WRA is run with $Max = 300$ and $c = 1$. The two algorithms, MGA and WRA, output loop cutsets of the same size in only 17 graphs and when the algorithms disagree, then in 95% of these graphs WRA performed better than MGA.

The actual run time of $WRA(G, 1, 300)$ is about 300 times slower than GA (or MGA) on $G$. On the largest random graph we used, it took 4.5 minutes. Most of the time is spend in the last improvement of WRA. Considerable run time can be saved by letting $Max = 5$. For all 700 graphs, WRA(G,1,5) has already obtained a better loop cutset than MGA. The largest improvement, with $Max = 300$, was from a weight of 58.0 ($log_2$ scale) to a weight of 35.9. The improvements in this case were obtained in iterations 1, 2, 36, 83, 189 with respective weights of 46.7, 38.8, 37.5, 37.3, 35.9 and respective sizes of 22, 18, 17, 18, and 17 nodes. On the average, after 300 iterations, the improvement for the larger 100 graphs was from a weight of 52 to 39 and from size 22 to 20. The improvement for the smaller 600 graphs was from a weight of 15 to 12.2 and from size 9 to 6.7.

The second experiment compared between GA, MGA and WRA on four real Bayesian networks showing that WRA outperformed both GA and MGA after a single call to SINGLEWGUESSI. The weight of the output continued to decrease logarithmically with the number of iterations. We report the results with $Max = 1000$ and $c = 1$. Run time was between 3 minutes for Water and 15 minutes for Munin1 on a Pentium 133 with 32M RAM. The results also indicate that Pearl's conditioning algorithm can not run on these graphs due

| Name | $|V|$ | $|E|$ | $|F^*|$ | GA | MGA | WRA |
|---|---|---|---|---|---|---|
| Water | 32 | 123 | 16 | 40.7 | 42.7 | 29.5 |
| Mildew | 35 | 80 | 14 | 48.1 | 40.5 | 39.3 |
| Barley | 48 | 126 | 20 | 72.1 | 76.3 | 57.3 |
| Munin1 | 189 | 366 | 59 | 159.4 | 167.5 | 122.6 |

Figure 3: Log size (base 2) of the loop cutsets found by MGA or WRA.

to the large cutset needed.

## Acknowledgement

We thank Seffi Naor for fruitful discussions.